# The Annotation Guideline of LST20 Corpus

**Prachya Boonkwan   Vorapon Luantangsrisuk   Sitthaa Phaholphinyo
Kanyanat Kriengket   Dhanon Leenoi   Charun Phrombut
Monthika Boriboon   Krit Kosawat   Thepchai Supnithi**

Language and Semantic Technology Lab (LST)
National Electronics and Computer Technology Center
112 Phahonyothin Road
Khlong Nueng, Khlong Luang District
Pathumthani 12120, Thailand

## Abstract

This report presents the annotation guideline for LST20, a large-scale corpus with multiple layers of linguistic annotation for Thai language processing. Our guideline consists of five layers of linguistic annotation: word segmentation, POS tagging, named entities, clause boundaries, and sentence boundaries. The dataset complies to the CoNLL-2003-style format for ease of use. LST20 Corpus offers five layers of linguistic annotation as aforementioned. At a large scale, it consists of 3,164,864 words, 288,020 named entities, 248,962 clauses, and 74,180 sentences, while it is annotated with 16 distinct POS tags. All 3,745 documents are also annotated with 15 news genres. Regarding its sheer size, this dataset is considered large enough for developing joint neural models for NLP. With the existence of this publicly available corpus, Thai has become a linguistically rich language for the first time.

# Contents

# Chapter 1

# Word Segmentation Guideline

In the LST20 Guideline, word segmentation generally abides by the formation of compound words described in the Inter-BEST 2009 Guideline. The rule of thumb is quite simple:

> A compound word is formed only if free morphemes are combined to form a new semantic concept that discards their original meanings.

However, a few exceptions allow the formation of compound words that merely retain the original meaning of free morphemes, which we shall explain along the way.

The annotation format for word segmentation is as follows. We assume that the data is in a raw format that preserves white spaces as shown below.

(1) อย่างไรก็ตามบริษัท เอบีซี จำกัดจะรีบแจ้งเตือนลูกค้าถึงปัญหาที่เกิดขึ้นทันที
'However, ABC Corporation will notify all customers about the current issues immediately.'

We annotate all word boundaries with a vertical bar '|' and preserve all white spaces as is. In this paper, we denote any white space with the square-cup symbol '⊔'.

(2) อย่างไรก็ตาม | บริษัท | ⊔ | เอบีซี | ⊔ | จำกัด | จะ | รีบ | แจ้ง | เตือน | ลูกค้า | ถึง | ปัญหา | ที่ | เกิด | ขึ้น | ทันที |
'However, ABC Corporation will notify all customers about the current issues immediately.'

The underlying assumption of word segmentation is that as many free morphemes should be delimited as possible, except that compound words must be formed so that the sentence conveys the right meaning. The following subsections describe the criteria of compound word formation and some exceptions.

## 1.1 Compound Word Formation

Any compound word is comprised of free morphemes, where their combination does not retain their original meanings. Some compound words are obvious and context-independent. For instance:

- แมวน้ำ [mæː.náːm] *sea lion* consists of two free morphemes: แมว [mæː] *cat* and น้ำ [náːm] *water*.

- กินใจ [kin.tɕaɪ] *be touching* consists of two free morphemes: กิน [kin] *eat* and ใจ [tɕaɪ] *heart*.

The other compound words are less clear and their formation depends on context interpretation, such as มีอายุ [miː.ʔaːjú] *elderly* vs. มี [miː] *have* + อายุ [ʔaːjú] *age*.

(3) a. pʰûː.tɕʰaːɪ kʰon níː duː   <u>miː.ʔaːjú</u>
       man    CL   this look elderly
       'This man looks elderly.'

   b. dèk  nák-rian    <u>miː</u>  ʔaːjú sìp.sɔ̌ːŋ piː
      child NOMZ-study have age  twelve CL.YEAR
      'The young student is 12 years old.'

## 1.2 Plant and Animal Names

Any specific plant and animal name is treated as a compound word. For example, these plant and animal names are treated as compound words.

- มะม่วงน้ำดอกไม้ [mámuâŋ.ná:m.dɔ̀:kmá:ɪ] *barracuda mango* consists of three free morphemes: มะม่วง [mámuâŋ] *mango*, น้ำ [ná:m] *water*, and ดอกไม้ [dɔ̀:kmá:ɪ] *flower*.

- เพลี้ยกระโดดสีน้ำตาล [pʰliá.kràdò:t.sǐ:.námta:n] *brown planthopper* consists of four free morphemes: เพลี้ย [pʰliá] *spittle bug*, กระโดด [kràdò:t] *jump*, สี [sǐ:] *color*, and น้ำตาล [námta:n] *brown*.

However, any part of plants and animals will be treated separately; for example, ขา | หลัง | เพลี้ยกระโดดสีน้ำตาล *hind leg of brown planthopper*.

(4)  kʰǎ:  lǎŋ   pʰliá.kràdò:t.sǐ:.námta:n
    leg   hind  brown planthopper
    'a hind leg of a brown planthopper'

## 1.3 Rhyming

Free morphemes with semantic or phonetic rhyming are also licensed to form a compound word. Semantically rhyming morphemes are combined to form a compound word. For example:

- ทรัพย์สิน [sáp.sǐn] *possession* is composed of two free morphemes: ทรัพย์ [sáp] *asset* and สิน [sǐn] *money*.

- เสาะแสวงหา [sɔ̀.sʷǎ:ŋ.hǎ:] *seek* consists of three free morphemes: เสาะ [sɔ̀], แสวง [sʷǎ:ŋ], หา [hǎ:], all of which meaning '*seek*'.

A morpheme can also be combined with another phonetic rhyming one to form a compound word. For example:

- โครมคราม [kʰro:m.kʰra:m] *smack! smack!* consists of two phonetically rhyming parts: โครม [kʰro:m] *smack!* and คราม [kʰra:m] *indigo*. The second part, whose meaning is discarded, is added to rhyme with the first morpheme.

- สะบักสะบอม [sàbàk.sàbɔm] *be badly bruised* consists of two phonetically rhyming parts: สะบัก [sàbàk] *shoulder blade* and สะบอม [sàbɔm] (meaningless). The second part is added to rhyme with the first morpheme.

- กระป๋งกระเป๋า [kràpǒŋ.kràpǎʊ] *bag* (intensified) consists of two phonetically rhyming parts: กระป๋ง [kràpǒŋ] (meaningless) and กระเป๋า [kràpǎʊ] *bag*. The first part is added to rhyme with the second morpheme.

Note that the meaning of the rhyming part is entirely discarded.

## 1.4 Reduplicatives

Words formed with reduplication are treated either morphologically or orthographically. In the case of morphological reduplication, words are explicitly repeated, perhaps with minor tonal and stress change; e.g. เด็กเด็ก [dèk.dèk] *children* (pluralized), แด๊งแดง [ˈdǽ:ŋ.dæ:ŋ] *red* (intensified), and แดงแดง [dæŋ.ˈdæ:ŋ] *reddish* (moderated). These morphological reduplicatives are treated as compound words.

On the other hand, orthographical reduplicatives, where punctuation mark 'ๆ' *mai yamok* is used, are treated separately. If *mai yamok* is used to modify the core morpheme, it is treated as a separate word, such as เด็ก | ๆ [dèk.dèk] *children* (pluralized), and เล็ก | ๆ *very small* (intensified). Otherwise, if *mai yamok* becomes a part of the word, e.g. ต่างๆ นานา [tà:ŋ.tà:ŋ.na:na:] *various* (intensified), and ทั่วๆ ไป [tʰuâ.tʰuâ.paɪ] *general* (moderated), it will be treated as a compound word.

## 1.5 Proverbs

Aphorisms, proverbs, and sayings are treated as compound words and annotated with respect to their syntactic functions. For example: ตาเป็นมัน [ta:.pen.man] (*staring*) *attentively*.

(5) nɔ́ːŋ     mɔːŋ kʰànǒm taː.pen.man ləːɪ
    younger brother look  snack   attentively   EMPHASIS
    'My younger brother stares at the snack very attentively.'

In the above example, the aphorism ตา เป็น มัน [taː.pen.man] can be literally translated as *the eyes are shimmering*.

## 1.6 Loanwords, Pseudo-Loanwords, and Courteous Terms

Loanwords from languages other than Thai are always treated as compound words. Most loanwords in Thai are from Pali and Sanskrit. For example, ชีววิทยา [tɕʰiːʋáʋíttʰája:] *Biology* is derived from Sanskrit words *jīva + vidyā > jīvavidyā*. Pseudo-loanwords are a combination of Thai and foreign words. For example, ราชวัง [râːttɕʰáʋaŋ] *royal palace* is a combination of Sanskrit word ราช [râːt] *king < rāja* and Thai word วัง [ʋaŋ] *palace*.

Courteous terms are a special vocabulary used when addressing the King, the Queen, royal family members, and Buddhist monks. These terms are also treated as compound words. For example, ฉลองพระเนตร [tɕʰàlɔ̌ːŋ.pʰrá.nêːt] *eyeglasses* is used in the court instead of แว่นตา [ʋɛ̂ntaː] *eyeglasses*. However, when a non-courteous term is combined with a courteous one, they are treated as separate words. For example, ถ้วย | พระสุธารส [tʰûaɪ.pʰrásùtʰaːrót] *tea cup* is composed of two words: ถ้วย [tʰûaɪ] *cup* and พระสุธารส [pʰrásùtʰaːrót] *tea*.

## 1.7 Prefixes

Despite its name, all kinds of prefixes are separated from the core part because they can generatively combine with lengthy phrases. These prefixes include: nominalizers (การ [kaːn] *action*, ความ [kʰwaːm] *abstract concept*, ผู้ [pʰûː] *person*, ชาว [tɕʰaːʋ] *citizen*, and นัก [nák] *professional*), adjectivizers (น่า [nâː] *likely*), adverbializers (โดย [doːɪ] *with*, and อย่าง [jàːŋ] *fashion*), courteous verbalizer ทรง [soŋ], and derivational prefixes (การ [kaːn] *action*, ชาว [tɕʰaːʋ] *citizen*, and นัก [nák] *professional*).

For the ease of understanding and typesetting, we will omit the separation of prefixes from the stem in some linguistic examples, if the stem consists of only one word. Instead of fully displaying the prefixes and stem in example 6a, we will reduce them into one chunk delimited by '-' as shown in example 6b.

(6) a. [NP kʰwaːm/FX [ADJP nâː/FX rák ]] kʰɔ̌ːŋ tʰəː     tɕʰâːŋ sàdùt.taː
       NOMZ        ADJZ  love   of  3RD.SING.FEM quite  be eye-catching
       'Her cuteness is quite eye-catching.'

    b. kʰwaːm-nâː-rák     kʰɔ̌ːŋ tʰəː     tɕʰâːŋ sàdùt.taː
       NOMZ-ADJZ-love  of    3RD.SING.FEM quite  be eye-catching
       'Her cuteness is quite eye-catching.'

## 1.8 Connectors and Prepositions

Thai allows multiple connectors and prepositions to juxtapose in the sentence. They will be segmented and annotated separately. For example, 'ไขควง อยู่ ใน ที่ บน ชั้น วาง ของ' *The screwdriver is on the shelf* consists of three consecutive prepositions, all of which being separated.

(7) kʰǎɪkʰuaŋ   jùː  naɪ  tʰîː  bon tɕʰán ʋaːŋ kʰɔ̌ːŋ
    screwdriver be  in   at   on  level  lay  thing
    'The screwdriver is on the shelf.'

Furthermore, 'หรือ แม้แต่ เขา ก็ อ่าน หนังสือ' *Even he also revises the lessons* consists of two consecutive connectors.

(8) rɯ̌ː  mɛ́ːtɛ̀ː  kʰǎʋ          kɔ̂ː  ʔàːn nǎŋsɯ̌ː
    or   even   3RD.SING.MASC also read book
    'Even he also revises the lessons.'

## 1.9 Punctuation Marks

All punctuation marks, including ๆ *mai yamok* (reduplication), ฯ *paiyal noi* (abbreviation), and ฯลฯ *paiyal yai* (et cetera), are treated as separate words. Consecutive non-Thai punctuation marks are treated a single token. URLs are also treated as single tokens.

# Chapter 2

# POS Tagging Guideline

In the LST20 Guideline, all Thai words are generally classified, according to their semantic contents, into two classes: content words and function words. The content words are then divided into nouns, verbs, adjectives, and adverbs. Meanwhile, the function words are divided into auxiliary, connector, classifier, prefix, interjection, negator, number, preposition, punctuation, and others. In total, there are 16 distinct POS tags as shown in Table 2.1.

The annotation format for POS tags is as follows. We assume that each sentence is annotated with word boundaries with respect to our word segmentation guideline, where each word is delimited with a vertical bar '|'. For example:

(9) อย่างไรก็ตาม | บริษัท | ␣ | เอบีซี | ␣ | จำกัด | จะ | รีบ | แจ้ง | เตือน | ลูกค้า | ถึง | ปัญหา | ที่ | เกิด | ขึ้น | ทันที |
'However, ABC Corporation will notify all customers about the current issues immediately.'

POS tags will be annotated to each word separated by a forward slash '/'. Therefore the above sentence will be annotated with POS tags as follows.

(10) อย่างไรก็ตาม/CC | บริษัท/NN | ␣/PU | เอบีซี/NN | ␣/PU | จำกัด/VV | จะ/AX | รีบ/VV | แจ้ง/VV | เตือน/VV | ลูกค้า/NN | ถึง/PS | ปัญหา/NN | ที่/CC | เกิด/VV | ขึ้น/AV | ทันที/AV |
'However, ABC Corporation will notify all customers about the current issues immediately.'

In the case of annotation ambiguity, we first classify a word by its semantic content. If it contributes to the meaning of the sentence in which it occurs, it is a content word, which we will further classify it with distributional test frames. Otherwise, if it rather denotes grammatical relationships between content words, we consider it a function word, which we will classify it based on the grammatical relationship it manifests.

## 2.1 Content Words

The content words are divided into four categories: noun, verb, adjective, and adverb. The definition of each tag is based on those defined in Thai Grammar [1]. We discern each of these categories with a simple set of distributional test frames, previously explored in [2].

**Noun** (`NN`) is a word used to identify any of a class of people, places, things, or abstract concepts (common noun), or to name a particular one of these (proper noun). We use the following distributional test frames to validate if a word is a noun.

**NN.1:** It can perform as the subject of a verb: ___ `VV AV`, *and*

**NN.2:** It can perform as the object of a verb: `NN VV` ___ `AV`, *and*

**NN.3:** It can perform as the complement of a preposition: `NN VV PS` ___ `AV`, *and*

**NN.4:** It can be modified by a classifier and an adjective: ___ `CL AJ`.

Here the underline ___ is a placeholder for a word to be tested, and each pair of parentheses denote an optional part in the test frames. If any word passes *all* of these test frames, it is said to be a noun.

For example, สุนัข [sùnák] *dog* is a noun because it passes the following test frames.

Table 2.1: 16 POS tags of LST20

| Tags | Names | Descriptions |
|------|-------|--------------|
| AJ | Adjective | A word naming an attribute, added to or grammatically related to a noun to modify or describe it |
| AV | Adverb | A word that modifies or qualifies an adjective, verb, or other adverb or a word group, expressing a relation of place, time, circumstance, manner, cause, degree etc. |
| AX | Auxiliary | A word used in forming the tenses, aspects, moods, and voices of the verbs or used in expressing necessity or possiblity |
| CC | Connector | A word used to connect clauses or sentences or to coordinate words in the same clause (conjunction), and a word that refers to an expressed or implied antecedent and attaches a subordinate clause to it (relative pronoun) |
| CL | Classifier | A word that indicates the semantic class or measurement unit to which a noun or an action belongs |
| FX | Prefix | A word placed before a noun, a noun phrase, a verb, or a verb phrase to adjust or qualify its meaning |
| IJ | Interjection | A word used for exclamation |
| NG | Negator | A word expressing negation |
| NN | Noun | A word used to identify any of a class of people, places, things, or abstract concepts (common noun), or to name a particular one of these (proper noun) |
| NU | Number | An arithmetical value expressed by a word, symbol, or figure, representing a particular quantity and used in counting and calculations and for showing an order in the series or for identification |
| PA | Particle | A word used with a phrase or a sentence used for linguistic nuance e.g. politeness, intention, belief, and question |
| PR | Pronoun | A word that refers either to a noun phrase or to an element in the discourse |
| PS | Preposition | A word governing a noun phrase or pronoun and expressing a relation to another word or element in the clause |
| PU | Punctuation | A mark used in writing to separate sentences and their elements and to clarify meaning |
| VV | Verb | A word used to describe an action, state, or occurrence, and forming the main part of the predicate of a sentence |
| XX | Others | A word having an ambiguous grammatical function or belonging to an unknown category |

(11) a. <u>sùnák/NN</u> ʋîŋ tɕúːt
   dog  run blazingly
   'The dog runs blazingly.'

 b. mɛ̂ː tiː <u>sùnák/NN</u> ʔìːk
   mother hit dog  again
   'Mother hits the dog again.'

 c. màt kràdòːt tɕàːk <u>sùnák/NN</u> ʔìːk
   flea jump from dog  again
   'The fleas jump from the dog again.'

 d. <u>sùnák/NN</u> tua tɔ̀ːpaɪ
   dog  CL next
   'the next dog'

**Verb** (VV) is a word used to describe an action, state, or occurrence, and forming the main part of the predicate of a sentence. We employ the following distributional test frames to validate if a word is a verb.

**VV.1:** It takes a subject: NN [$_{VP}$ AX ___ ] AV, *or*

**VV.2:** It takes either an object or a noun complement: [$_{VP}$ AX ___ NN ] AV, *or*

**VV.3:** It takes direct and indirect objects: [$_{VP}$ AX ___ NN$_{DO}$ NN$_{IO}$ ] AV, *or*

**VV.4:** It takes a topic and a property: NN$_T$ [$_{VP}$ AX NN$_P$ ___ (NN) ] AV, *or*

**VV.5:** It requires a complementing verb phrase after it: NN [$_{VP}$ ___ VV (NN) ] AV.

**VV.6:** *And* it is the complement of a relative pronoun: NN VV NN CC VP.

Note that the test frames VV.1 to VV.5 describe five kinds of verbs: intransitive verb, transitive verb, ditransitive verb, incorporative verb, and linking verb, respectively. The test frame VV.6 checks if the word can be used as the complement of a relative pronoun. If any word passes *any* of VV.1-VV.5 and it also passes VV.6, it is said to be a verb.

For example, บิน [bin] *fly*, กิน [kin] *eat*, ให้ [hâɪ] *give*, คล้าย [kʰláːɪ] *be similar*, กรุณา [kàrúnaː] *be kind*, and อ้วน [ʔûan] *be plump* are verbs because they pass at least one of test frames VV.1-VV.5 and also passes VV.6.

(12) a. nók [$_{VP}$ tɕà <u>bin/VV</u> ] nɛ̂ː.nɛ̂ː
   bird  FUT fly  surely
   'The bird will fly surely.' (VV.1)

 b. pʰîː  tɕàp nók tʰîː [$_{VP}$ tɕà <u>bin/VV</u> ]
   older brother catch bird RELPRO FUT fly
   'My younger brother catches the bird that will fly away.' (VV.6)

(13) a. [$_{VP}$ tɕà <u>kin/VV</u> kʰâːʊ ] nɛ̂ː.nɛ̂ː
    FUT eat  rice surely
   'I will eat rice surely.' (VV.2)

 b. pʰîː  tɕàp nók tʰîː [$_{VP}$ tɕà <u>kin/VV</u> kʰâːʊ ]
   older brother catch bird RELPRO FUT eat  rice
   'My older brother catches birds that would eat rice.' (VV.6)

(14) a. [$_{VP}$ kamlaŋ <u>hâɪ/VV</u> ŋən nɔ́ːŋ  ] nɛ̂ː.nɛ̂ː
    CONT  give money younger brother surely
   'I am giving my younger brother some money surely.' (VV.3)

 b. mɛ̂ː  kliàt jǐŋ  tʰîː [$_{VP}$ kamlaŋ <u>hâɪ/VV</u> ŋən nɔ́ːŋ  ]
   mother hate woman RELPRO  PROG give money younger brother
   'Mother hates the woman that is giving my brother some money.' (VV.6)

(15) a. pʰîː      [_VP_ kʰəːɪ nâː.taː <u>kʰláːɪ/VV</u> pʰɔ̂ː ] maː.kɔ̀ːn  
       older brother     PAST face   be similar father   previously  
       'My older brother used to have a face similar to my father.' (VV.4)

     b. mɛ̂ː mi: pʰîː tʰîː [_VP_ kʰəːɪ nâː.taː <u>kʰláːɪ/VV</u> pʰɔ̂ː ]  
       mother have older brother RELPRO    PAST face   be similar father  
       'Mother has an older brother that had a face similar to her father.' (VV.6)

(16) a. nák-rian     [_VP_ <u>kàrúnaː/VV</u> faŋ   kʰruː ] diː.diː  
       NOMZ-study    be kind    listen teacher   well  
       'Students, please listen to the teacher well.' (VV.5)

     b. mɛ̂ː tɕʰɔ̂ːp nák-rian     tʰîː [_VP_ <u>kàrúnaː/VV</u> faŋ   kʰruː ]  
       mother like   NOMZ-study RELPRO    be kind    listen teacher  
       'Mother likes the students that kindly listen to the teacher.' (VV.6)

(17) a. sùnák [_VP_ kʰəːɪ <u>ʔuân/VV</u> ] maː.kɔ̀ːn  
       dog      PAST be plump    previously  
       'The dog used to be plump.' (VV.1)

     b. mɛ̂ː mi: sùnák tʰîː [_VP_ kʰəːɪ <u>ʔuân/VV</u> ]  
       mother have dog    RELPRO    PAST be plump  
       'Mother has a dog that used to be plump.' (VV.6)

Note that อ้วน [ʔuân] *be plump* is considered a verb in Thai, although its equivalent is an adjective in English. This kind of verbs belong to a special class called *attributive verbs*, where they describe an attribute or a quality of the subject. Thai relative pronouns are omittable if the context is clear as shown in example 18.

(18) mɛ̂ː mi: [_NP_ sùnák (tʰîː) <u>ʔuân/VV</u> ]  
      mother have    dog    RELPRO be plump  
      'Mother has a plump dog.'

In this case, we still take into account อ้วน [ʔuân] *be plump* as a verb because it is the complement of an omitted relative pronoun. There are in fact only a handful of genuine Thai adjectives as we shall see below.

     **Adjective** (AJ) is a word naming an attribute, added to or grammatically related to a noun to modify or describe it. Our distributional test frames for an adjective are as follows.

**AJ.1:** It shows an attribute or definiteness of the modified noun: NN (CL) ___ VV, *or*

**AJ.2:** It quantifies the modified noun that follows: ___ NN VV, *or*

**AJ.3:** It quantifies the number and classifier that follow: NN ___ NU CL VV, *or*

**AJ.4:** It quantifies the circumjacent number and classifier: NN NU ___ CL VV.

If it passes *any* of these test frames, it is said to be an adjective.

     For example, ต่อไป [tɔ̀ːpaɪ] *next*, บาง [baːŋ] *some*, เกือบ [kɯ̀ə:p] *almost*, and กว่า [kʷàː] *more than* are adjectives because they pass one of these test frames.

(19) a. pʰɛ̌ːn ʔan <u>tɔ̀ːpaɪ/AJ</u> jɔ̂ːtjiâm  
       plan CL next     be excellent  
       'The next plan is excellent.' (AJ.1)

     b. <u>baːŋ/AJ</u> roːŋ.rian pʰàːn  
       some    school   pass  
       'Some schools pass the test.' (AJ.2)

     c. rót <u>kɯ̀ə:p/AJ</u> sìp kʰan sǐa  
       car almost     ten CL    be broken  
       'Almost ten cars are broken.' (AJ.3)

d. rót sìp k<sup>w</sup>à:/AJ k<sup>h</sup>an sǐă
car ten more than CL be broken
'More than ten cars are broken.' (AJ.4)

**Adverb** (AV) is a word that modifies or qualifies an adjective, verb, or other adverb or a word group, expressing a relation of place, time, circumstance, manner, cause, degree etc. The following are the distributional test frames for an adverb.

**AV.1:** It modifies or qualifies the preceding verb phrase: NN VV (NN) ___, *or*

**AV.2:** It modifies the succeeding sentence, making it a question: ___ NN VV NN, *or*

**AV.3:** It qualifies the succeeding sentence (e.g. *in fact*): ___ NN VV NN, *or*

**AV.4:** It resembles a verb that qualifies the preceding verb with direction (e.g. ไป [paɪ] *go*, มา [ma:] *come*, ขึ้น [k<sup>h</sup>ûn] *ascend*, and ลง [loŋ] *descend*), motion (e.g. เข้า [k<sup>h</sup>âʊ] *enter*, ออก [ʔɔ̀:k] *exit*, เร็ว [reʊ] *be fast*, and ช้า [tɕ<sup>h</sup>á:] *be slow*), acceleration (e.g. เข้า [k<sup>h</sup>âʊ] *hurry up*), anticipation (e.g. ดู [du:] *look forward* and ออก [ʔɔ̀:k] *expect*), asking (e.g. ไว้ [ʊáɪ] *continue* and เสีย [sǐă] *lose*), disappointment and disagreement (e.g. เสีย [sǐă] *lose*), decision (e.g. เสีย [sǐă] *lose*), and causing (e.g. ให้ [hâɪ] *cause*): NN VV NN ___.

If it passes *any* of these test frames, it is said to be an adverb.

For example, ซก [sôk] *soaking*, ทำไม [t<sup>h</sup>ammaɪ] *why*, ที่จริง t<sup>h</sup>î:tɕiŋ *in fact*, and เข้า [k<sup>h</sup>âʊ] *hurry up* are adverbs because they pass one of these test frames.

(20) a. sɯ̂ɔ: pìăk sôk/AV
shirt wet soaking
'The shirt is soaking wet.' (AV.1)

b. t<sup>h</sup>ammaɪ/AV sɯ̂ɔ: lɔ́ sǐ:
why shirt be stained color
'Why is the shirt stained with colors?' (AV.2)

c. t<sup>h</sup>î:tɕiŋ/AV p<sup>h</sup>ɔ̂: mi: ŋən
in fact father have money
'In fact, father has some money.' (AV.3)

d. nák-rian t<sup>h</sup>am ka:n.bâ:n k<sup>h</sup>âʊ/AV
NOMZ-study do homework hurriedly
'Students, do your homework hurriedly!' (AV.4)

## 2.2 Function Words

Function words are divided into 12 categories: auxiliary, connector, classifier, prefix, interjection, negator, number, particle, pronoun, preposition, punctuation, and others.

**Auxiliary** (AX) is a word used in forming the tenses, aspects, moods, and voices of the verbs or used in expressing necessity or possiblity. Table 2.2 lists some auxiliary words found in LST20 Corpus. Among those, the passive and causative voices are expressed by specific constructions in examples 21 and 22, respectively.

(21) a. nák-rian t<sup>h</sup>ù:k/AX k<sup>h</sup>ru: t<sup>h</sup>am.t<sup>h</sup>ô:t
NOMZ-study PASS teacher punish
'The student was punished by the teacher.'

b. nák-rian t<sup>h</sup>ù:k/AX k<sup>h</sup>ru: ríp ka:tu:n
NOMZ-study PASS teacher confiscate comic book
'The comic book was confiscated from the student by the teacher.'

(22) a. k<sup>h</sup>ru: hâɪ/AX nák-rian ʔà:n năŋsɯ̌:
teacher CAUSE NOMZ-study read book
'The teacher asks the students to read the book.'

Table 2.2: List of some auxiliary words found in LST20 Corpus

| Auxiliary (AX) | Tense | Aspect | Mood | Voice |
|---|---|---|---|---|
| กำลัง [kamlaŋ] | | continuous | | |
| คง [kʰoŋ] | | | hypothetical | |
| ควร [kʰuan] | | | imperative | |
| ค่อย [kʰɔ̂ɪ] | | | cohortative | |
| เคย [kʰəːɪ] | past | habitual | | |
| จะ [tɕà] | future | | | |
| จง [tɕoŋ] | | | imperative | |
| จวน [tɕuan] | near-future | | | |
| ได้ [dâɪ] | | perfective | | |
| ต้อง [tɔ̂ŋ] | | | imperative | |
| น่า [nâː] | | | potential | |
| ถูก [tʰùːk] | | | | passive |
| โดน [doːn] | | | | passive |
| เพิ่ง [pʰɤ̂ŋ] | | perfective | | |
| มัก [mák] | | habitual | | |
| ยัก [ják] | | | counterfactual | |
| ยัง [jaŋ] | | perfective | | |
| ย่อม [jɔ̂m] | | habitual | | |
| แล้ว [lɛ́ːʊ] | past | | | |
| ไว้ [ʋáɪ] | | continuous | imperative | |
| เสร็จ [sèt] | | perfective | | |
| ให้ [hâɪ] | | | | causative |
| ทำให้ [tʰamhâɪ] | | | | causative |
| อยู่ [jùː] | | continuous | | |
| อยู่แล้ว [jùːlɛ́ːʊ] | | optative | | |

   b. kʰruː bɔ̀ːk nák-rian <u>hâɪ/AX</u> ʔàːn năŋsɯ̌ː
      teacher tell NOMZ-study CAUSE read book
      'The teacher tells the students to read the book.'

**Connector** (CC) is a word used to connect clauses or sentences or to coordinate words in the same clause (conjunction), and a word that refers to an expressed or implied antecedent and attaches a subordinate clause to it (relative pronoun). There are three kinds of Thai connectors as follows.

1. **General conjunction:** Connectors of this kind conjoin more than one constituent of the same types, forming a coordinate structure. They are further divided into three subtypes: single conjunctions (e.g. และ [lɛ́] *and*, ก็ [kɔ̂ː] *also*), correlative conjunctions (e.g. ถ้า…ก็ [tʰâː kɔ̂ː] *if…then*, เพราะ…เลย [pʰrɔ́ ləːɪ] *because…so*), and subordinate conjunctions (e.g. เช่น [tɕʰên] *for example*, เพราะ [pʰrɔ́] *because*).

2. **Cohesive marker:** Connectors of this kind conjoin two complete sentences, forming a discourse relationship. Some cohesive markers include แต่ทว่า [tɛ̀ːtʰáʋâː] *nevertheless*, ในที่สุด [naɪtʰîːsùt] *finally*, and อย่างไรก็ตาม [jàːŋraɪkɔ̂ːtaːm] *however*.

3. **Relative pronoun:** Connectors of this kind modifies the preceding noun phrase or sentence with a subordinate clause. Some relative pronouns include ซึ่ง [sɯ̂ŋ] *that/which*, ที่ [tʰîː] *that/which*, and ผู้ [pʰûː] *who/whom*.

Note that more than one connector are allowed to occur in one sentence. For example, three connectors occur in this sentence.

(23) <u>sɯ̂ŋ/CC</u> kʰăʊ tʰùːk tɕàp lăːɪ kʰráŋ <u>tɛ̀ː/CC</u> <u>kɔ̂ː/CC</u> mâɪ kʰèt
    <u>RELPRO</u> he PASS arrest many time but still NEG be terrified
    '…which he was arrested many times, but he is still not terrified.'

The distinction between connectors and prepositions is not always clear due to homographs. One prominent feature that discerns them is the use of negator. For example, ระหว่าง [ráʋàːŋ] *while* is either a preposition or a connector, depending on the context. In example 24, it is considered a connector because it connects two sentences, therefore licensing the use of negator.

(24) a. pʰɔ̂: ʋa:ŋ.pʰɛ̌:n tʰîaʋ ráʋà:ŋ/CC rápprátʰa:n a:hǎ:n jen
father plan travel while eat meal evening
'Father is planning his trip while he is having the dinner.'

b. pʰɔ̂: ʋa:ŋ.pʰɛ̌:n tʰîaʋ ráʋà:ŋ/CC mâɪ rápprátʰa:n a:hǎ:n jen
father plan travel while NEG eat meal evening
'Father is planning his trip while he is not having the dinner.'

It is, on the other hand, considered a preposition in example 25, because it connects a sentence and a noun phrase, disallowing the use of negator.

(25) a. pʰɔ̂: ʋa:ŋ.pʰɛ̌:n tʰîaʋ ráʋà:ŋ/PS ʋe:la: a:hǎ:n jen
father plan travel while time meal evening
'Father plans his trip during the dinner.'

b. *pʰɔ̂: ʋa:ŋ.pʰɛ̌:n tʰîaʋ ráʋà:ŋ/PS mâɪ ʋe:la: a:hǎ:n jen
father plan travel while NEG time meal evening

**Classifier** (CL) is a word that indicates the semantic class or measurement unit to which a noun or an action belongs. The classifiers are divided into four subtypes: unit classifier (e.g. ตัว [tua] *body*, คัน kʰan *car*), collective classifier (e.g. โหล [lǒ:] *dozen*, เครือ [kʰrɯa:] *vine*), measurement classifier (e.g. เมตร [mé:t] *meter*, หลา [lǎ:] *yard*), and frequency classifier (e.g. ครั้ง [kʰráŋ] *time*, ฟอด [fɔ̂:t] *kissing time*).

Classifiers are used for counting and specifying a noun or an action. Example 26 illustrates how unit classifier tua *body* is used to form adjectival phrases for specifying the core noun and counting it, respectively. Example 27 shows two usages of collective classifiers: ฝูง [fǔ:ŋ] *swarm of bees* preceding the core noun and โขลง [kʰlǒ:ŋ] *herd of elephants* succeeding it, respectively. In example 28, frequency classifiers โครม [kʰro:m] *crashing time* and ครั้ง [kʰráŋ] *time* are used to specify the manner and count the frequency of the action, respectively.

(26) mi: sùnák [ADJP tua/CL jài ] [ADJP sǎ:m tua/CL ]
EXIST dog CL.ANIMAL big three CL.ANIMAL
'There are three big dogs.'

(27) [NP fǔ:ŋ/CL pʰɯ̂ŋ ] tɕo:mti: tɕʰá:ŋ.pà: [ADJP tʰáŋ kʰlǒ:ŋ/CL ]
COL.SWARM bee attack wild elephant entire COL.HERD
'A swarm of bees attack the entire herd of wild elephants.'

(28) troŋ.ní: rót.faɪ kʰə.ɪ tɕʰon [ADVP kʰro:m/CL jài ] [ADVP sǎ:m kʰráŋ/CL ]
here train PAST crash CL.CRASH big three CL.TIME
'The trains loudly crashed here three times.'

**Prefix** (FX) is a word placed before a noun, a noun phrase, a verb, or a verb phrase to adjust or qualify its meaning. Prefixes are divided into two subtypes: inflectional prefix and derivational prefix. Inflectional prefixes include: nominalizers (การ [ka:n] *action*, ความ [kʰwa:m] *abstract concept*, ผู้ [pʰû:] *person*, ชาว [tɕʰa:ʋ] *citizen*, and นัก [nák] *professional*), adjectivizers (น่า [nâ:] *likely*), adverbializers (โดย [do:ɪ] *with*, and อย่าง [jà:ŋ] *fashion*), and courteous verbalizer ทรง [soŋ]. Derivational prefixes include การ [ka:n] *action*, ชาว [tɕʰa:ʋ] *citizen*, and นัก [nák] *professional*.

Note that multiple prefixes may be used to form a complex constituent. In example 29, adjectivizer น่า [nâ:] *likely* combines with รัก [rák] *love*, forming น่ารัก [nâ:.rák] *cute*. Then nominalizer ความ [kʰwa:m] combines with น่ารัก [nâ:rák], forming ความน่ารัก [kʰwa:m.nâ:.rák] *cuteness*. Example 30 shows how adverbializer อย่าง [jà:ŋ] combines with verb phrase มี ชั้นเชิง [mi: tɕʰántɕʰə:ŋ] *have tactics*, forming an adverbial phrase modifying verb เต้นรำ [tên.ram] *dance*. Then nominalizer การ [ka:n] combines with such verb phrase to form a noun phrase.

(29) [NP kʰwa:m/FX [ADJP nâ:/FX rák ]] kʰɔ̌:ŋ tʰə: tɕʰâ:ŋ sàdùt.ta:
NOMZ ADJZ love of 3RD.SING.FEM quite be eye-catching
'Her cuteness is quite eye-catching.'

(30) kʰruː     sɔ̌ːn  [NP ka:n/FX [VP tên.ram [ADVP jàːŋ/FX [VP miː   tɕʰán.tɕʰəːŋ ]]]]
     teacher teach        NOMZ        dance           ADVZ            have tactics
     'The teacher teaches dancing with tactics.'

**Interjection** (IJ) is a word used for exclamation. Words of this category express an emotion such as pain (อุย [ʔuːɪ] *ouch!*), joy (ไชโย [tɕʰaɪjoː] *hurrah!*), surprise (โอ้โห [ʔôːhǒː] *wow!*), shock (ว้าย [ʋáːɪ] *gosh!*), disappointment (อ้าว [ʔâːʊ] *oh!*), and regret (โถ่เอ๊ย [tʰôː.ʔə́ːɪ] *oh no!*).

**Negator** (NG) is a word expressing negation. For example, example 31 shows that negator ไม่ [mâɪ] can occur before a verb, while negator ก็หาไม่ [kɔ̂ː.hǎː.mâɪ] occurs after a verb. However, correlative negators หา…ไม่ [hǎː mâɪ] is also used in Classical Thai (circa 15th Century), as shown in example 32.

(31) a. mɛ̂ː    mâɪ/NG  jùː  bâːn
        mother NEG    stay home
        'Mother is not home.'

     b. nɔ́ːŋ              dəːn mâɪ/NG ráʋaŋ  rót
        younger brother walk NEG    be aware car
        'My younger brother walks unaware of the cars.'

     c. kʰaʊ            tɕà  sǎmnɯ́k kɔ̂ː.hǎː.mâɪ/NG
        3RD.SING.MASC FUT regret    NEG
        'He does not regret at all.'

(32) ʔɔːtɕâʊ    hǎː/NG tôŋ  paɪ kiam rɯːa  hâɪ kʰâː   mâɪ/NG
     2ND.SING NEG   should go  prepare row boat for 1ST.SING NEG
     'Thou dost not have to prepare the row boat for me.'

**Number** (NU) is an arithmetical value expressed by a word, symbol, or figure, representing a particular quantity and used in counting and calculations and for showing an order in the series or for identification. For any cardinal number, it can be written in any numeral format, e.g. 123.45, or spelt out as words, e.g. สามสิบสอง [sǎːm sìp sɔ̌ːŋ] *thirty two* (lit. *three ten two*). For any ordinal number, it can be written in any numeral format or spelt out as order words, e.g. แรก [rɛ̂ːk] *first*, กลาง [klaːŋ] *middle*, and สุดท้าย [sùt.tʰáːɪ] *last*. Order markers such as ที่ [tʰîː], if apparent, will be annotated as a preposition. For example:

(33) rót kʰɔ̌ːŋ tɕʰǎn       kʰâʊ pen tʰîː/PS sùt.tʰáːɪ/NU
     car of   1ST.SING arrive be  at       last
     'My car arrives in the last place.'

**Particle** (PA) is a word used with a phrase or a sentence used for linguistic nuance e.g. politeness, intention, belief, and question. Politeness particles include ครับ [kʰráp] *masculine politeness*, ค่ะ [kʰâ] *feminine politeness*, and วะ [ʋá] *impoliteness*. Intention particles include นะ [ná] *invitation* and *confirmation*, เนี่ย [niâ] *emphasis*, and เถิด [tʰə̀ːt] *asking*. Belief particles include สินะ [sì.ná] *likelihood*, ซิ [sí] *certainty*, and มั้ง [máŋ] *uncertainty*. Question particles include ใช่ไหม [tɕʰâɪ.mǎɪ] *yes/no question*, ยัง [jaŋ] *yet*, and หรือเปล่า [rɯ̌ː.plàʊ] *yes/no question*.

It is sometimes challenging to distinguish between particles and adverbs, because both of them modify sentences. One prominent trait of particles is they can also modify other kinds of phrases, such as noun phrases and preposition phrases, while the adverbs cannot. For example, particle นะ [ná] modifies both a sentence and a noun phrase in example 34. Meanwhile, แล้ว [lɛ́ːʊ] is considered an auxiliary, because it modifies only sentences but cannot modify any noun phrases as shown in example 35.

(34) a. raʊ     paɪ pʰáttʰája: kan     ná/PA
        1ST.PLU go  Pattaya     together INVITE
        'Let's go to Pattaya together.'

     b. pʰáttʰája: ná/PA
        Pattaya    CONFIRM
        '(Let's go to) Pattaya then.'

(35) a. raʊ     paɪ pʰáttʰája: kan     lɛ́ːʊ/AX
        1ST.PLU go  Pattaya     together PAST
        'We went to Pattaya already.'

    b. \* pʰáttʰája: <u>lǽːʊ/AX</u>
       Pattaya   PAST
       —

**Pronoun** (PR) is a word that refers either to a noun phrase or to an element in the discourse. The pronouns are divided into four subtypes: personal pronouns (e.g. กระผม [kràpʰŏm] 1ST.SING.MASC.POLITE, มึง [mɯŋ] 2ND.IMPOLITE), demonstrative pronouns (e.g. นี่ [nîː] *this/these*, นั่น [nân] *that/those*), interrogative and indefinite pronouns (e.g. อะไร [ʔàraɪ] *what*, ใครๆ [kʰraɪ.kʰraɪ] *everyone*), and partitive pronouns (e.g. กัน [kan] *each other*, ต่าง [tàːŋ] *each*).

Note that kinship terms and professional terms can be used to refer to persons as if they are pronouns. For example, นักเรียน [nák-rian] *student* refers to the listener who is a student in the following sentence.

(36) <u>nák-rian</u>  tɕà  paɪ năɪ  kʰá
     NOMZ-study FUT go where POLITE
     'Where are you going, student?'

However, they will rather be annotated as nouns than as pronouns, because they pass all test frames of nouns NN.1-NN.4. In contrast, pronouns do not pass test frame NN.4; for instance:

(37) a. <u>nák-rian/NN</u>  kʰon  tɔ̀ːpaɪ
      NOMZ-study CL.PERS next
      'The next student'

     b. \* <u>kʰăʊ/PR</u>     kʰon  tɔ̀ːpaɪ
       3RD.SING.MASC CL.PERS next
       —

**Preposition** (PS) is a word governing a noun phrase or pronoun and expressing a relation to another word or element in the clause. Prepositions are divided into four categories: location (e.g. บน [bon] *on*, กลาง [klaːŋ] *middle*), comparison (e.g. กว่า [kʷàː] *than*, เท่า [tʰâʊ] *as*, อย่างกับ [jàːŋ.kàp] *like*), instrument (e.g. ด้วย [duaɪ] *with*, ทาง [tʰaːŋ] *by*), and exemplification (e.g. เช่น [tɕʰên] *for example*, ได้แก่ [dâɪkæ̀ː] *namely*).

**Punctuation** (PU) is a mark used in writing to separate sentences and their elements and to clarify meaning. This category includes all English and Thai punctuation marks, e.g. exclamation mark, question mark, ๆ *mai yamok* (reduplication mark), ฯ *paiyal noi* (abbreviation), and ฯลฯ *paiyal yai* (et cetera).

**Others** (XX) is a word having an ambiguous grammatical function or belonging to an unknown category. We annotate the remaining ambiguities taking place in our corpus for further studies.

# Chapter 3

# Named Entity Annotation Guideline

In the LST20 annotation guideline, named entities are annotated in the corpus. Listed in table 3.1, ten types of named entities are chosen because they are beneficial for general NLP applications, such as sentiment analysis, information extraction, automatic summarization, and social media monitoring.

Our annotation format of named entities complies with the `BIEO` tagging convention. The boundary of each named entity is annotated with the prefixes `B_`, `I_`, and `E_`, denoting the beginning, the intermediate, and the ending, respectively, while the tag `O` denotes the outside of a named entity. We assume that each sentence is annotated with word boundaries and POS tags with respect to our guideline. That means each word is delimited with a vertical bar '|' and POS tags are annotated to each word separated by a forward slash '/'. For example,

(38) อย่างไรก็ตาม/CC | บริษัท/NN | ␣/PU | เอบีซี/NN | ␣/PU | จำกัด/VV | จะ/AX | รีบ/VV | แจ้ง/VV | เตือน/VV | ลูกค้า/NN | ถึง/PS | ปัญหา/NN | ที่/CC | เกิด/VV | ขึ้น/AV | ทันที/AV |

Named entities will be annotated to each word separated by a forward slash '/'. Therefore the above sentence will be annotated with named entities as follows.

(39) อย่างไรก็ตาม/CC/O | บริษัท/NN/B_ORG | ␣/PU/I_ORG | เอบีซี/NN/I_ORG | ␣/PU/I_ORG | จำกัด/VV/E_ORG | จะ/AX/O | รีบ/VV/O | แจ้ง/VV/O | เตือน/VV/O | ลูกค้า/NN/O | ถึง/PS/O | ปัญหา/NN/O | ที่/CC/O | เกิด/VV/O | ขึ้น/AV/O | ทันที/AV/O |

In the above example, the annotated named entity is an organization's name, which is annotated with tag `ORG`.

The named entity tags are generally divided into three groups: personal entity, collective entity, and referential entity. Personal entities include person name, title, and designator. Collective entities include organization, location, and brand name. Finally, referential entities include date and time, measurement unit, number, and terminology.

## 3.1 Personal Entity

Personal entity is a text chunk that refers to or specifies a person name or a family name. It can be decomposed into three components: person name, title, and designator. If a personal entity consists of more than one component, each of them will be tagged separately.

**Person name** (PER): This part is the core of the personal entity and does not include a title, a profession, and an order of family members. For example, consider the personal entity 'นายกรัฐมนตรี ดร. มหาธีร์ บิน โมฮัมหมัด' *Prime Minister Doctor Mahathir bin Mohamad*.

(40) na:jók.rátt$^h$àmontri: dóktɔ̂: máhǎ:t$^h$i: bin mo:hammàt
    prime minister      doctor Mahathir bin Mohammad
    'Prime Minister Doctor Mahathir bin Mohamad'

Only the core มหาธีร์ บิน โมฮัมหมัด *Mahathir bin Mohamad* will be annotated with the PER tag.

(41) นายกรัฐมนตรี | ดร. | มหาธีร์/B_PER | บิน/I_PER | โมฮัมหมัด/E_PER |

**Title** (TTL): This part is a kinship term (e.g. พี่ [p$^h$î:] *older brother/sister*, น้อง [nɔ́:ŋ] *younger brother/sister*) or a social status (e.g. นาย [na:ɪ] *Mister*, ดร. [dóktɔ̂:] *Doctor*) of a person entity. Only the social status ดร. *Doctor* will be annotated with the TTL tag.

Table 3.1: Named entity tagset

| Tags | Names | Descriptions |
|------|-------|--------------|
| TTL | Title | Family relationship, social relationship, and permanent title |
| DES | Designation | Position and professional title |
| PER | Person | Name of a person or family |
| ORG | Organization | Name of organization, office, or company |
| LOC | Location | Name of a land according to geo-political borders (e.g. city, province, country, international regions, and oceans) |
| DTM | Date and time | Time or a specific period of time |
| BRN | Brand | Name of brand, product, and trademarks |
| MEA | Measurement | Measurement unit and quantity of things |
| NUM | Number | Number specifying the quantity as a part of measurement unit |
| TRM | Terminology | Domain-specific words |

(42) นายกรัฐมนตรี | ดร./B_TTL | มหาธีร์ | บิน | โมฮัมหมัด |

**Designator** (DES): This part indicates a professional title (e.g. อาจารย์ [ʔa:tɕa:n] *teacher*, พระ [pʰrá] *monk*), a political position (e.g. ประธานาธิบดี [pràtʰa:na:tʰíbadi:] *President*, นายกรัฐมนตรี [na:jók.rátʰàmontri:] *Prime Minister*), a rewarded title (e.g. แชมเปี้ยน [tɕʰɛ̌mpiân] *champion*, เดอะสตาร์ [dɔ̀.sta:] *The Star*), or an academic title (e.g. ศาสตราจารย์ [sà:ttra:tɕa:n] *Professor*, สารวัตร [sǎ:rávát] *Police Inspector*). Only the political position นายกรัฐมนตรี *Prime Minister* will be annotated with the DES tag.

(43) นายกรัฐมนตรี/B_DES | ดร. | มหาธีร์ | บิน | โมฮัมหมัด |

## 3.2 Collective Entity

Collective entity is a text chunk that refers to an organization of people, to a location in which people live, or to a brand name indicating an organization.

**Organization** (ORG): This kind of collective entities refer to an organization in which people work together, such as government, parties, councils, offices, unions, companies, sport teams, and coalitional organizations. For example, consider the text 'สำนักงาน ปลัดกระทรวง ␣ กระทรวง วิทยาศาสตร์ และ เทคโนโลยี' *Office of Permanent Secretary, Ministry of Science and Technology*.

(44) sǎmnák.ŋa:n pàlàt.kràsuaŋ    kràsuaŋ ʋíttʰája:sà:t lɛ́ tʰékno:lo:ji:
     office       permanent secretary ministry science      and technology
     'Office of Permanent Secretary, Ministry of Science and Technology'

This text consists of two organization names: one is an office, and the other one is a ministry that governs the office. However, they will be annotated separately as two named entities.

(45) สำนักงาน/B_ORG | ปลัดกระทรวง/E_ORG | ␣ | กระทรวง/B_ORG | วิทยาศาสตร์/I_ORG | และ/I_ORG | เทคโนโลยี/E_ORG |

**Location** (LOC): This kind of collective entities specify a geographical location, a construction, a facility, or a natural terrain in which people live or work in, such as continents, cities, house addresses, buildings, bridges, and waterfalls. For example, consider the text 'ที่ โรงแรม อินโดจีน ␣ อำเภอ อรัญประเทศ ␣ จังหวัด สระแก้ว' *at Indo-China Hotel, Aranyaprathet County, Sakaew Province*.

(46) tʰî: ro:ŋ.rɛ:m ʔindotɕi:n  ʔampʰə: ʔàranjápràtʰêt tɕaŋʋàt sàkɛ̂:ʋ
     at  hotel    Indo-China  county  Aranyaprathet   province Sakaew
     'at Indo-China Hotel, Aranyaprathet County, Sakaew Province'

This text consists of three location names: one is a hotel, the next one is a county, and the last one is a province. Each of them will be annotated separately as three named entities.

(47) ที่ | โรงแรม/B_LOC | อินโดจีน/E_LOC | ␣ | อำเภอ/B_LOC | อรัญประเทศ/E_LOC | ␣ | จังหวัด/B_LOC | สระแก้ว/E_LOC |

**Brand name** (BRN): This kind of collective entities refer to brands, products, or trademarks. For example, consider the text 'ไก่ ทอด เคเอฟซี' *KFC fried chicken*.

(48) kàɪ   tʰɔ̂:t  kʰeɪ.ʔéf.siː
     chicken fry K-F-C
     'KFC fried chicken'

In this text, we annotate only the brand name in the text, resulting in the following annotation.

(49) ไก่ | ทอด | เคเอฟซี/B_BRN |

The distinction between organization, location, and brand name is sometimes unclear. One prominent feature of the organization names is that they can perform some activities and take some effects as if they are a group of people, while the others are referred to as a position or a product, respectively. For example, these texts contain the same names 'ภาควิชา ภาษาศาสตร์' *Department of Linguistics*.

(50) a. ʔaːtɕaːn sɔ̌ːn  tʰîː  pʰâːk.ʋítɕʰa: pʰaːsǎːsàːt  kʰráp
        lecturer teach at  department  linguistics  POLITE
        'I (a lecturer) work for Department of Linguistics.'

   b. pʰâːk.ʋítɕʰa: pʰaːsǎːsàːt jùː tʰîː tɕʰán hòk
      department  linguistics  be at floor six
      'Department of Linguistics is on the sixth floor.'

Such name in example 50a is considered as an organization, in which a group of lecturers work together. On the other hand, the name in example 50b is rather taken into account as a location. Therefore, we annotate such names with ORG and LOC, respectively.

(51) a. อาจารย์ | สอน | ที่ | ภาควิชา/B_ORG | ภาษาศาสตร์/E_ORG | ครับ

   b. ภาควิชา/B_LOC | ภาษาศาสตร์/E_LOC | อยู่ | ที่ | ชั้น | หก |

## 3.3 Referential Entity

Referential entity is a text chunk that refers to date and time, to measurement unit, to number, or to terminology.

**Date and time** (DTM): This kind of entities refer to a specific date and time and a duration, such as seasons, public holidays, festivals, and names of ages. For example, consider the following text 'โรงเรียน เปิด ใน เดือน พฤษภาคม ของ ทุก ปี' *School starts in May of every year*.

(52) roːŋ.rian  pɤ̀ːt  naɪ  dɯən  pʰrɯ́tsàpʰaːkʰom  kʰɔ̌ːŋ  tʰúk  piː
     school    open  in   month  May               of     every year
     'School starts in May of every year.'

In this case, we will annotate the specific time with the DTM tag.

(53) โรงเรียน | เปิด | ใน | เดือน/B_DTM | พฤษภาคม/I_DTM | ของ/I_DTM | ทุก/I_DTM | ปี/E_DTM |

Note that the DTM tag is very specific to a particular point or duration of time. If a reference of time or duration is indeterminate, we will not annotate it with this tag. For example, these references of time will not be annotated with the DTM tag.

(54) a. pʰǒm         jùː  piː  sɔ̌ːŋ kʰráp
        1ST.SING.MASC be   year two  POLITE
        'I am in the second year.'

   b. tɕɤː kan  tɔːn  tʰîaŋ  náʔ
      meet RECIP at   midday INVITE
      'Let's meet each other at midday.'

**Measurement unit** (MEA): This kind of entities specify a measurement unit, a percentage, ratio, quantity, and capacity without any quantifier. The classifier must be a standard measurement unit, not any other general one. For example, consider the following text 'ไอโฟน SE ราคา เริ่มต้น ประมาณ 15,000 บาท' *The starting price of iPhone SE is approximately 15,000 Baht*.

(55) ʔaɪfoːn ʔésʔiː raːkʰaː rə̂ːmtôn pràmaːn nɯ̀ŋ.mɯ̀ːn.hâː.pʰan bàːt
iPhone SE price start approximately fifteen thousand Baht
'The starting price of iPhone SE is approximately 15,000 Baht.'

In this case, we will annotate the currency unit with the MEA tag. Note that the quantifier ประมาณ [pràmaːn] *approximately* is not included.

(56) ไอโฟน SE ราคา เริ่มต้น ประมาณ 15,000/B_MEA บาท/E_MEA

**Number** (NUM): This kind of entities refer to a number and a range without any quantifier, when the classifier is not a measurement unit. For example, consider the following text 'ปี นี้ มี นักศึกษา ประมาณ 3,000 คน' *There are approximately 3,000 students this year*.

(57) piː níː miː nák.sɯ̀ksǎː pràmaːn sǎːmpʰan kʰon
year this EXIST student approximately three thousand CL
'There are approximately 3,000 students this year.'

In this case, we annotate only the number 3,000 with the NUM tag. Note that the quantifier ประมาณ [pràmaːn] *approximately* and the classifier คน [kʰon] *person* are not annotated.

(58) ปี | นี้ | มี | นักศึกษา | ประมาณ | 3,000/B_NUM | คน |

**Terminology** (TRM): This kind of entities refer to domain-specific words, such as financial terms, scientific terms, and political terms. For example, consider the following text 'ไวรัส โควิด-19 แพร่ ระบาด ไป ทั่ว โลก' *COVID-19 has spread worldwide*.

(59) ʋaɪrás kʰoːʋìt sìpkâːʋ pʰrɛ̂ː rábàːt paɪ tʰûâ lôːk
virus COVID-19 propagate spread go through World
'COVID-19 has spread worldwide.'

In this case, we annotate only the terminology, not any other general terms, with the TRM tag. Note that the word ไวรัส [ʋaɪrás] *virus* is not annotated.

(60) ไวรัส | โควิด-19/B_TRM | แพร่ | ระบาด | ไป | ทั่ว | โลก |

# Chapter 4

# Clause Segmentation Guideline

In the LST20 Guideline, texts are divided into clauses. Our annotation format for clause boundaries follows the BIEO tagging convention. The boundary of each clause is annotated with B_CLS (beginning), I_CLS (intermediate), and E_CLS (ending), respectively. The outside of the clauses is annotated with the O tag. We assume that each text is annotated with word boundaries, POS tags, and named entities with respect to our guideline. That means each word is delimited with a verticle bar '|', while POS tags and named entities are annotated to each word separated by a forward slash '/'. For example,

(61) น.พ./NN/B_PER|จรัล/NN/E_PER|⊔|กล่าว/VV/O|ต่อ/AV/O|ว่า/CC/O|⊔|จาก/PS/O|การ/FX/O|สอบสวน/VV/O|โรค/NN/O|พบ/VV/O|ว่า/CC/O|⊔|ผู้/NN/O|ที่/CC/O|เสีย/VV/O|ชีวิต/NN/O|ก่อนหน้า/PS/O|นี้/PR/O|⊔|มี/VV/O|ประวัติ/NN/O|สัมผัส/VV/O|ไก่/NN/O|ที่/CC/O|ตาย/VV/O|ด้วย/PS/O|โรค/NN/B_TRM|ไข้หวัด/NN/I_TRM|นก/NN/E_TRM|
'Dr. Charan says the disease investigation shows that the deceased previously had contact with chickens infected with avian influenza.'

Clause boundaries will be annotated to each word separated by a forward slash '/'. Therefore, the above text will be annotated with clause boundaries as follows.

(62) น.พ./NN/B_PER/B_CLS|จรัล/NN/E_PER/I_CLS|⊔/PU/O/I_CLS|กล่าว/VV/O/I_CLS|ต่อ/AV/O/I_CLS|ว่า/CC/O/E_CLS|⊔|จาก/PS/O/B_CLS|การ/FX/O/I_CLS|สอบสวน/ VV/O/I_CLS|โรค/NN/O/I_CLS|พบ/VV/O/I_CLS|ว่า/CC/O/E_CLS|⊔|ผู้/NN/O/B_CLS|ที่/CC/O/I_CLS|เสีย/VV/O/I_CLS|ชีวิต/NN/O/I_CLS|ก่อนหน้า/PS/O/I_CLS|นี้/PR/O/ I_CLS|⊔/PU/O/I_CLS|มี/VV/O/I_CLS|ประวัติ/NN/O/I_CLS|สัมผัส/VV/O/I_CLS|ไก่/NN/O/E_CLS|ที่/CC/O/B_CLS|ตาย/VV/O/I_CLS|ด้วย/PS/O/I_CLS|โรค/NN/B_TRM/ I_CLS|ไข้หวัด/NN/I_TRM/I_CLS|นก/NN/E_TRM/E_CLS|
'Dr. Charan says the disease investigation shows that the deceased previously had contact with chickens infected with avian influenza.'

Note that the first word of each clause is underlined. White spaces are sometimes incorporated to a clause and, in this case, they will be annotated with the punctuation mark tag (PU).

Since there is no definite consensus for clause boundaries in Thai, the choice of clause segmentation is arbitrarily personal. In this paper, we define the notion of *clause* to be a part of the sentence that contains at least one verb. If a clause contains either an explicit subject and a predicate, or only a predicate part whose verb does not require any syntactic subject (e.g. มี [miː] *exist*), it is said to be an *independent clause*. Otherwise, if it does not contain any syntactic subject that the predicate part requires, it is then said to be a *dependent clause* (also known as *subordinate clause*). We also define the clause marker to be:

- **Subordinate connector:** e.g. ซึ่ง [sûŋ] *that/which* (relative pronoun), ถ้า [tʰâː] *if*, and ว่า [ʋâː] *that* (subordinate conjunction),

- **Cohesive marker:** e.g. อย่างไรก็ตาม [jàːŋraikɔ̂ːtaːm] *however*, and นอกจากนี้ [nɔ̂ːkteàːkniː] *in addition*,

- **List marker:** e.g. เช่น [tɕʰên] *for example*, ได้แก่ [dâɪkæ̀ː] *namely*, and ตามลำดับ [taːm.lamdàp] *respectively*.

- **Particle:** e.g. ครับ [kʰráp] *masculine politeness*, and นะ [ná] *invitation*, or

- **Question adverb:** e.g. อย่างไร [jàːŋraɪ] *how*, and ไหม [mǎɪ] *yes/no question*.

We allow any clause to be preceded and followed by one or more clause markers. For instance, clause markers are underlined in the following examples.

(63) a. naːɪ.pʰɛ̂ːt tɕàran klàːʊ tɔ̀ː   <u>ʊâː/CC</u>
        doctor   Charan say   continue that
        'Dr. Charan says that …'

   b. <u>tʰîː/CC</u> taːɪ duâɪ rôːk   kʰâɪ.ʊàt nók
       that    die with disease flu      bird
       '…that die of avian influenza'

   c. <u>nɔ̂ːktɕàːkníː/AV</u> jaŋ miː     ʔìːk lǎːɪ  kɔːrániː
       in addition        still EXIST yet   many  case
       'In addition, there are yet more cases …'

Clues for clause boundaries are all syntactic and computationally deterministic in most cases. The following clues are used to determine the beginning or the end of a clause.

**Paragraph boundary:** Any paragraph boundary marks the end of the clause and the beginning of the next one.

**White space:** If two adjacent text chunks, each containing verbs, are delimited by white spaces, and at least one clause marker occurs right before or next to the delimiter, then we can mark the separation of the clauses here. For instance, two clauses in example 64 are separated by a white space ␣.

(64) a. จาก การ สืบสวน โรค พบ <u>ว่า/CC</u> ␣ ผู้ ป่วย เคย สัมผัส ไก่ ติด เชื้อ
       'From the disease investigation, it was found that patients had contact with infected chickens.'

   b. tɕàk kaːn    sɔ̀ːpsuǎn  rôːk   pʰóp <u>ʊâː/CC</u>
       from NOMZ investigate disease find  that
       'From the disease investigation, it was found that …'

   c. pʰûː-puàɪ      kʰɤːɪ sǎmpʰàt kàɪ      tìt     tɕʰɯ́ː
       NOMZ-be ill PAST touch    chicken be infected germ
       '…patients had contact with infected chickens.'

**Clause marker:** If a text chunk contains a subordinate connector or a relative pronoun, it will mark the beginning of the next clause. For example, consider the sentence 'ฉัน ไม่ ทราบ <u>ว่า</u> ทำไม เขา ไม่ แถลง ข่าว' *I do not know why he did not call a press conference*.

(65) [<sub>CLS</sub> tɕʰǎn    mâɪ sâːp  ] [<sub>CLS</sub> <u>ʊâː/CC</u> tʰammaɪ kʰǎʊ mâɪ  tʰàlɛ̌ːŋ   kʰàːʊ ]
          1ST.SING NEG know            that    why      he    NEG announce news
     'I do not know why he did not call a press conference.'

Subordinate connector ว่า [ʊâː] *that* marks the beginning of the second clause, although there is no white space in the text chunk.

# Chapter 5

# Sentence Segmentation Guideline

In the LST20 Guideline, texts are also divided into sentences. The boundary of each sentence is annotated by appending an empty symbol '‖' after it. For example,

(66) เขา/PR/O/B_CLS|ก็/CC/O/I_CLS|ไม่/NG/O/I_CLS|ได้/AX/O/I_CLS|โทร/VV/O/I_CLS|มา/AV/O/I_CLS|คุย/VV/O/E_CLS | ␣ | ต่าง/AJ/O/B_CLS | คน/CL/O/I_CLS | ก็/CC/O/ I_CLS | ต่าง/AJ/O/I_CLS | อยู่/VV/O/I_CLS | กัน/AV/O/I_CLS | ไป/AV/O/E_CLS | ␣ | ดิฉัน/PR/O/B_CLS | คิด/VV/O/I_CLS | แล้ว/AV/O/I_CLS | ว่า/CC/O/E_CLS | ␣ | ควร/AX/O/ B_CLS | วางตัว/VV/O/I_CLS | อย่างไร/AV/O/I_CLS | และ/CC/O/I_CLS | ควร/AX/O/I_CLS | ทำ/VV/O/I_CLS | อะไร/PR/O/I_CLS | ต่อไป/AV/O/E_CLS | ␣ |

'He did not call me to settle the issue. We just stayed in our corners. I have considered where my position is and what my next move should be.'

Sentence boundaries are annotated by the empty symbol '‖'. Therefore, the above sentence will be annotated with sentence boundaries as follows.

(67) a. เขา/PR/O/B_CLS|ก็/CC/O/I_CLS|ไม่/NG/O/I_CLS|ได้/AX/O/I_CLS|โทร/VV/O/ I_CLS|มา/AV/O/I_CLS|คุย/VV/O/E_CLS ‖
'He did not call me to settle the issue.'

b. ต่าง/AJ/O/B_CLS|คน/CL/O/I_CLS|ก็/CC/O/I_CLS|ต่าง/AJ/O/I_CLS|อยู่/VV/O/ I_CLS|กัน/AV/O/I_CLS|ไป/AV/O/E_CLS ‖
'We just stayed in our corners.'

c. ดิฉัน/PR/O/ B_CLS|คิด/VV/O/I_CLS|แล้ว/AV/O/I_CLS|ว่า/CC/O/E_CLS|␣|ควร/ AX/O/B_CLS|วางตัว/VV/O/I_CLS |อย่างไร/AV/O/I_CLS|และ/CC/O/I_CLS|ควร/AX/ O/I_CLS|ทำ/VV/ O/I_CLS|อะไร/PR/O/I_CLS|ต่อไป/AV/O/E_CLS ‖
'I have considered where my position is and what my next move should be.'

Note that each sentence is stored as a separate line and the sentence boundaries are underlined.
    Similar to the clause boundaries, there is no definite consensus for sentence boundaries in Thai and the choice for sentence segmentation is arbitrarily personal. In this paper, we believe that sentence boundaries are ambiguously marked with the white spaces, whose other usages of them are English's comma and stylistics. However, unlike the clause boundaries, some clues for sentence boundaries require semantic interpretation. The following clues are used to determine the beginning or the end of a sentence.
    **Paragraph boundary:** Any paragraph boundary marks the end of the sentence.
    **Topic shift:** If a clause starts with a cohesive marker (for example, อย่างไรก็ตาม [jàːŋraikɔ̂ːtaːm] *however*, นอกจากนี้ [nɔ̂ːktɕàːkníː] *in addition*), such marker indicates the beginning of the next sentence.
    **Subject shift:** This criterion is mainly based on semantic interpretation. If two adjacent clauses do not share the same subjects (explicit or contextually implied), we mark the separation of the sentences here. Otherwise, we concatenate them into one sentence. For example, consider the following clauses.

(68) a. kʰăʊ   kɔ̂ː   mâɪ   dâːɪ   tʰroː   maː   kʰuɪ
       he    also  NEG   PERF   call    come  talk
       'He did not call me to settle the issue.'

    b. <u>tàːŋ</u> <u>kʰon</u> kɔ̂ː tàːŋ jùː kan paɪ
        each person also each stay RECIP go
        'We just stayed in our corners.'

These clauses are segmented as separate sentences, because they do not share the same subjects (as underlined). Otherwise, if they share the same subjects, they will be concatenated as one sentence. For example, these clauses are concatenated to form a single sentence, because they share the same subjects.

(69)  a. [ <u>tɕʰɯɤ́ː ʔétɕʰ.hâː.ʔen.nɯ̀ŋ</u> ]$_i$ pen tɕʰàpʰɔ́ naɪ sàt pìːk
        germ H5N1                    be only in animal wing
        'H5N1 spreads only among the avians …'

    b. $\phi_i$ pʰə̂ŋ rábàːt sùː kʰon mɯɤ̂ː tôn piː
        just spread to human at beginning year
        '…and (it) just spread to humans in the beginning of this year.'

The subject in the first clause is referred to by a zero anaphora in the second clause.

    **Direct speech:** The construction of direct speech consists of a reporting verb and a quote enclosed in a pair of parentheses. We will treat this construction as a large sentence in our guideline. For example, consider this sentence นายกรัฐมนตรี กล่าว ยืนยัน ว่า ␣ " ไม่ พบ ผู้ ป่วย ราย ใหม่ " *The Prime Minister confirms that "zero new patients have been found"*. We will treat this entire chunk as a sentence despite its length.

(70)  นายกรัฐมนตรี/NN/B_TTL/B_CLS | กล่าว/VV/O/I_CLS | ยืนยัน/VV/O/I_CLS | ว่า/CC/O/E_CLS | ␣ | "/PU/O/B_CLS | ไม่/NG/O/I_CLS | พบ/VV/O/I_CLS | ผู้/FX/O/I_CLS | ป่วย/ VV/O/I_CLS | ราย/CL/O/I_CLS | ใหม่/VV/O/I_CLS | "/PU/O/E_CLS ‖

    **Indirect speech:** The construction of indirect speech consists of a reporting verb, a subordinate conjunction, and one or more subordinate clauses. In our guideline, we treat the verb, the subordinate conjunction, and the first subordinate clauses, whose subjects are shared, as one sentence. The remaining subordinate clauses are annotated with sentence boundaries in the same fashion. For example, consider the following clauses. They will be concatenated to form a sentence, because the subordinate clauses share the same subjects.

(71)  a. <u>naːɪ.pʰɛ̂ːt tɕaran</u> klàːʊ <u>vâː</u>
        doctor Charan say that
        'Doctor Charan says that …'

    b. tɕʰɯɤ́ː ʔétɕʰ.hâː.ʔen.nɯ̀ŋ pen tɕʰàpʰɔ́ naɪ sàt pìːk
        germ H5N1                be only in animal wing
        'H5N1 spreads only among the avians …'

    c. pʰə̂ŋ rábàːt sùː kʰon mɯɤ̂ː tôn piː
        just spread to human at beginning year
        '…and it just spread to humans in the beginning of this year.'

    **Item list:** Any clause that starts with a list marker will be combined to the previous clause to form a sentence. For example, consider the sentence 'โรงงาน ของ เขา ผลิต เครื่อง ดื่ม หลาย อย่าง ␣ <u>เช่น</u> ␣ เบียร์ ␣ น้ำ ดื่ม ␣ ชา เขียว ␣ ฯลฯ' *His factory manufactures various kinds of drinks such as beers, drinking water, green tea, etc*.

(72)  a. roːŋŋaːn kʰɔ̌ːŋ kʰǎʊ pʰàlìt kʰrɯɤ̂ːŋ dɯ̀ːm lǎːɪ jàːŋ
        factory of 3RD.SING.MASC manufacture ware drink various kind
        'His factory manufactures various kinds of drinks …'

    b. <u>tɕʰên</u> bia náːm dɯ̀ːm tɕʰaː kʰǐɤʊ
        for example beer water drink tea green
        '…for example, beers, drinking water, green tea, etc.'

These clauses are combined to form a sentence.

    **Particle:** Since Thai particles always posit in the final position of the sentence, we can predetermine the sentence boundaries with them. For example, consider the following clauses.

(73) a. pʰrûŋníː tɕəː kan sìp moːŋ <u>ná</u>
      tomorrow meet RECIP ten hours INVITE
      'Tomorrow let's meet up at 10.00 hours.'

   b. tɕà dâːɪ miː ʋeːlaː triam ʔèːkkàsǎːn
      will ALLOW have time prepare document
      'We will have some time to prepare the documents.'

These clauses are treated as separate sentences, because the first clause ends with an invitation particle นะ [ná].

# Chapter 6

# Structure and Format

## 6.1 Genre Distribution

LST20 Corpus consists of 3,745 articles with the genre distribution illustrated in figure 6.1. The top-3 popular genres are politics, crime and accident, and economics, which is congruent with Thailand's political turmoil during that period of time.

## 6.2 Available Data Format

LST20 Corpus is available at `https://aiforthai.in.th` in the CoNLL-2003-style format. The latter is simply a tab-delimited text file containing four columns: word, POS tag, NE boundary, and clause boundary. Each sentence is delimited by an empty line. Boundary tags for named entities and clauses comply with the `BIEO` convention. A glimpse of the dataset is shown in figure 6.2.

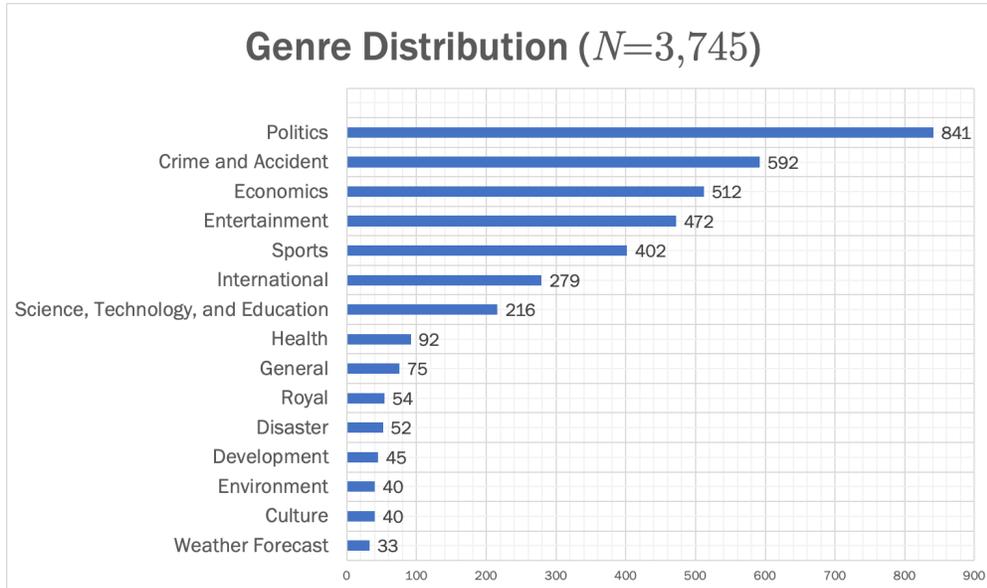

Figure 6.1: Genre distribution of LST20

Figure 6.2: A glimpse of LST20